\documentclass[letterpaper, 10 pt, conference]{ieeeconf}  

\IEEEoverridecommandlockouts                              
\usepackage{float}
\usepackage{graphicx}
\usepackage{placeins}
\usepackage{amsmath}
\usepackage{booktabs}
\usepackage[version=4]{mhchem}
\usepackage{siunitx}
\usepackage{longtable,tabularx}
\usepackage{tensor}
\usepackage{bm}
\usepackage{comment}
\usepackage{tikz}
\usepackage[font=footnotesize]{caption}
\usepackage{amsmath}
\usepackage[makeroom]{cancel}
\usepackage{multicol}

\usepackage{amsfonts}
\makeatletter
\def\widebreve{\mathpalette\wide@breve}
\def\wide@breve#1#2{\sbox\z@{$#1#2$}%
     \mathop{\vbox{\m@th\ialign{##\crcr
\kern0.08em\brevefill#1{1.0\wd\z@}\crcr\noalign{\nointerlineskip}%
                    $\hss#1#2\hss$\crcr}}}\limits}
\def\brevefill#1#2{$\m@th\sbox\tw@{$#1($}%
  \hss\resizebox{#2}{\wd\tw@}{\rotatebox[origin=c]{90}{\upshape(}}\hss$}
\makeatletter

\newcommand{\ffc}[3]{\tensor*{{#1}}{_{_{#2, #3}}}} 

\newcommand{\unitvec}[1]{\hat{\mathbf{#1}}}

\newcommand{\tm}{\mathbf{T}}

\newcommand{\trans}{\mathbf{p}}

\newcommand{\rotmat}{\mathbf{R}}

\newcommand{\unitrot}{\mathbf{\unitvec {w}}}
\newcommand{\taa}{\mathbf{\Upsilon}}

\newcommand{\axisangle}{\mathbf{r}}

\newcommand{\spaceframe}{\mathbf{O}}

\newcommand{\topframe}{\mathbf{T}}
\newcommand{\botframe}{\mathbf{B}}

\newcommand{\tj}{\mathbf{JiT}}
\newcommand{\bj}{\mathbf{JiB}}

\newcommand{\spaceToTopPlate}{\ffc{\tm}{\spaceframe}{\topframe}}
\newcommand{\spaceToBotPlate}{\ffc{\tm}{\spaceframe}{\botframe}}

\newcommand{\Li} {\mathbf{L}_{\mathbf{i}}}
\newcommand{\jointLengths}{\mathbf{L}_{\mathbf{1:6}}}

\newcommand{\spaceCoordsTopJoints}{\ffc{\trans}{\spaceframe}{\tj}}
\newcommand{\spaceCoordsBotJoints}{\ffc{\trans}{\spaceframe}{\bj}}

\newcommand{\IK}[2]{\mbox{IK}( {#1}, {#2} )}

\title{\LARGE \bf
Metrics and Optimization of Internal Poses for Highly Redundant Truss-Like Serialized Parallel Manipulators 
}

\author{William Chapin$^{1}$ and Erik Komendera$^{2}$
\thanks{$^{1}$William Chapin is an undergraduate student at Virginia Tech
        {\tt\small wchapin@vt.edu}}%
\thanks{$^{2}$Dr. Komendera is an assistant professor of Mechanical Engineering at Virginia Tech
        {\tt\small komendera@vt.edu}}%
}

\begin{document}

\maketitle
\thispagestyle{empty}
\pagestyle{empty}

\begin{abstract}

This paper presents a kinematic definition of a serialized Stewart platform designed for autonomous in-space assembly called an Assembler. The Assemblers architecture describes problems inherent to the inverse kinematics of over-actuated mixed kinematic systems. This paper also presents a methodology for optimizing poses. In order to accomplish this with the Assembler system, an algorithm for finding a feasible solution to its inverse kinematics was developed with a wrapper for a nonlinear optimization algorithm designed to minimize the magnitude of forces incurred by each actuator. A simulated version of an Assembler was placed into a number of representative poses, and the positions were optimized. The results of these optimizations are discussed in terms of actuator forces, reachability of the platform, and applicability to high-payload structure assembly capabilities.

\end{abstract}

\section{Introduction and Motivation}


Autonomous assembly carries the potential to revolutionize exploration of space by enabling the construction of larger spaceborne structures, reducing the risk and expense of using humans and increasing accessibility to space. Current surface assembly robots operate in precise, calibrated environments, and have regular access to human maintenance. In space, none of these assumptions can be made, so systems must be robust in their own right. Robots in space, especially deep space, must be expected to operate with a degree of autonomy, must be capable of detecting and correcting error, and must be able to actuate precisely.

Current paradigms for the deployment of on-orbit infrastructure require either monolithic launches within a single ascent vehicle, such as with the Hubble Space Telescope, or on-orbit assembly by Astronauts, whose time and presence is expensive and limited. Robotic construction by contrast allows for assembly of structures without humans in the loop. Autonomous assembly of truss structures has been demonstrated at NASA Langley Research Center (LaRC) with the Lightweight Surface Manipulation System (LSMS)\cite{lsms}, a large tendon-actuated serial arm, and NASA's Intelligent Jigging and Assembly Robot (NINJAR), a Stewart platform truss jig. The two robots worked in concert to demonstrate the assembly of a two bay truss structure \cite{wong2018validation}. Potential near-term applications include large communications antennae, as was proposed with Archinaut \cite{archinaut2017}, large assembled telescopes such as LUVOIR\cite{luvoir2017} and iSAT\cite{iSAT}, and radiation shields.

The modular Assembler architecture arose from the work presented first in \cite{wong2018validation} and \cite{iros2017}. It consists of two or more stacked Stewart platforms (SPs), each consisting of two plates connected by six linear actuators, enabling 6 degree of freedom (DOF) motion. The Assembler concept is of particular research interest to the space industry's desire to find alternatives to large, expensive, relatively flexible, single use-case serial manipulators for tasks that can only be performed in micro-gravity. 
The Assembler concept, by contrast, is cheap, easily mass produced, highly redundant, and corrects some of the instability and inaccuracy of long reach manipulators by virtue of its architecture as an actuatable truss structure. 
This paper defines a methodology to produce initial and optimal pose solutions for a four-SP Assembler stack \cite{moser_autonomous_2019} system in keeping with desired metrics such as force and plate angular deviation reduction. It expands on both the processes of simply generating valid kinematic poses for Assembler stacks and the methodology first presented in \cite{balaban_inverse_2018}. Instead of a two dimensional representation, this work accounts for the full kinematic freedom of the robot, aiming to allow trials on prototype hardware. 
\section{Definition of a Serialized Stewart Platform}
\subsection{Definition of Terms}
This work uses two representations for transformations and rotations in 3D Euclidean space. The first is the transformation matrix, $\tm$ (\ref{transStdFrm}). 
\begin{equation}
\label{transStdFrm}
    \tm = 
    \begin{bmatrix}
    \rotmat & \trans \\
    0 & 1
    \end{bmatrix}
\end{equation}
The upper left quadrant is the rotation matrix, denoted $\rotmat$, which is a 3x3 matrix. The upper right quadrant contains a 3x1 vector, denoted $\trans$, which defines the translation (meters, in this case) of the object assigned the matrix. The lower quadrants complete the 4x4 transformation matrix. While this method provides a natural way to perform transformations, it requires 12 values plus the bottom row to represent a 6 degree of freedom quantity, which requires enforcing 6 constraints in any solver.

The second method is the Translation-Axis-Angle (TAA) representation, which has 6 components for 6 degrees of freedom, consisting of three translational elements, and 3 orientation rotations around a standard axis, denoted $\taa$, as shown below. The rotation component is represented as an \emph{axis-angle} multiplying the rotation value by its unit rotation axis.

\begin{equation}
    \label{DefTaa}
    \taa =
    \begin{pmatrix}
        \trans \\
        \axisangle
    \end{pmatrix}
    =
    \begin{pmatrix}
        \trans \\
        \theta \unitrot
    \end{pmatrix}
\end{equation}

\subsection{Stewart Platforms}
A Stewart platform is a manipulation mechanism consisting of two plates joined together by six linear actuators which are anchored at three joint nodes on each plate. Frequently, the actuators of a Stewart platform are secured to each plate with a ball joint, but they can also be secured with universal joints, provided that the sum of degrees of freedom at each end of each actuator is at least 5. 

\begin{figure}[b]
    \centering
    \includegraphics[width=.49\linewidth]{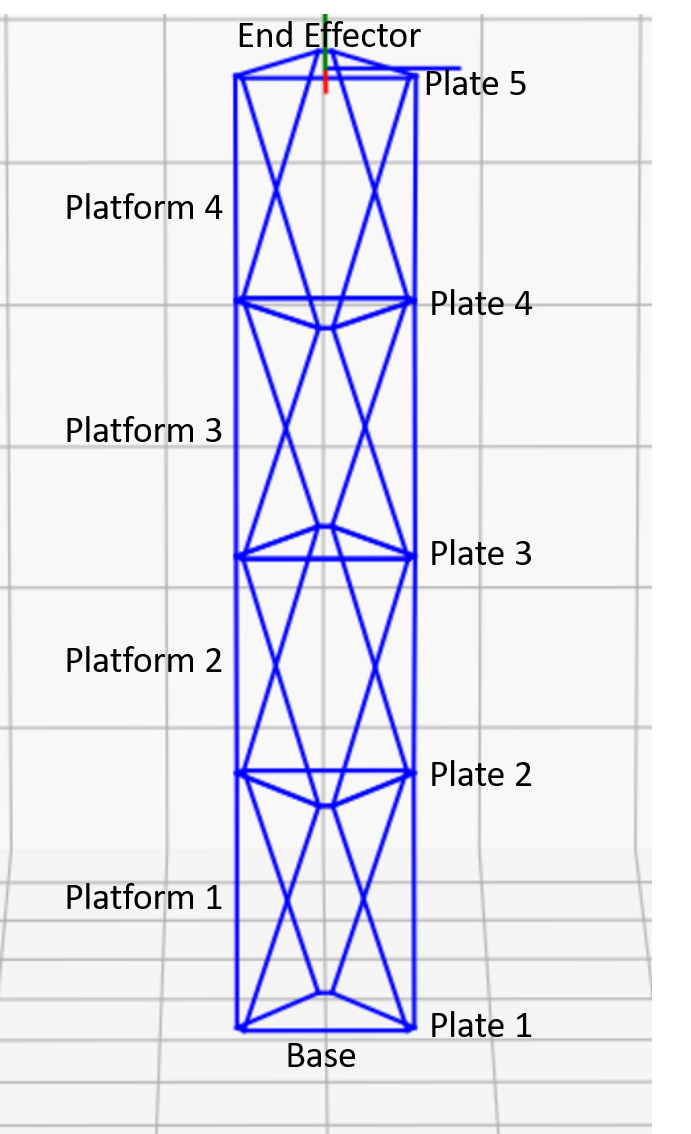}
    \includegraphics[width=.31\linewidth]{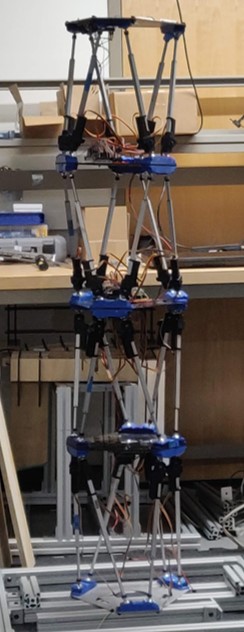}
    \caption{Left: Simulated Assembler with Plate and Platform Indices Labelled.
    Right: Physical Assembler Prototype}
    \label{fig:AssemblerDiagram}
\end{figure}

For each Stewart platform, the top plate transformation is defined by the matrix $\spaceToTopPlate$ and the bottom plate is defined by $\spaceToBotPlate$, both in the global frame $\spaceframe$. Each plate possesses a set of joint nodes, whose locations are linked to their respective transforms, and between each matching pair exists actuators denoted $\jointLengths$. These joint positions in the global frame are denoted $\spaceCoordsTopJoints$ and $\spaceCoordsBotJoints$ (top and bottom) respectively per joint $\mathbf{i}$, and do not carry rotational information. In Fig. \ref{fig:AssemblerDiagram}, the plates are denoted \emph{$P_i$}, with each individual Stewart platform denoted \emph{$SP_i$}.

The length of the legs, $\jointLengths$, can be determined through the use of a closed form inverse kinematic (IK) function, in the form $\IK{\spaceToBotPlate} {\spaceToTopPlate}$ (\ref{SPIK}).

\begin{equation}
    \label{SPIK}
    \Li = ||\spaceCoordsTopJoints - \spaceCoordsBotJoints|| 
\end{equation}

The forward kinematics (FK) of the Stewart platform are not consequential to an inverse kinematic, nonlinear optimization problem, so they are not described here. But except in specific cases, SP forward kinematics are solved numerically.

\subsection{Serial-Parallel Mechanisms}
The Assembler system, consisting of serialized Stewart platforms, can be abstractly represented as a serial arm. 
By treating the plates of the Assembler as the joints of a serial arm, the forward kinematic equation of the Assembler (\ref{AssemblerOFK}) closely matches the forward kinematics of a standard serial arm consisting of multiple revolte joints using the product of exponentials method (\ref{SerialFK}). 

\begin{equation}
    \label{AssemblerOFK}
    T_{ee} = T_{O} * \prod_{i=2}^{n}T_{P_n}
\end{equation}
\begin{equation}
    \label{SerialFK}
    T_{ee} = T_{O} * \prod_{i=1}^{n}e^{S_i\theta_i}
\end{equation}

While this approximation is valid given the assumption of using the plate positions only as inputs for the forward kinematic function, the actuator lengths must be calculated by inverse kinematics after the plate positions are determined. The direct formulation of forward kinematics for the Assembler takes the actuator lengths as inputs and returns an end-effector position. The plate positions for each actuator must be calculated via numerical approximation for each platform, and then applied via a serial forward kinematic function (\ref{AssemblerOFK}). Complicating this is the fact that, for any given set of Stewart platform actuator lengths, numerous positions may exist, which is multiplied by each plate.

Inverse kinematics of the Assembler can also be understood in the context of numerical serial arm inverse kinematics. Given a desired end effector pose, numerical approximation could be used to determine individual plate configurations, which could in turn be used to calculate actuator lengths of each constituent platform. 

This approach has two primary weaknesses:
\begin{itemize}
    \item Approximating an Assemblers plate as a serial arm joint fails because the 6DOF limits of its movement are dependent on the current configuration of the corresponding platform's legs, while serial arm joint limits are generally independent of the configuration of other joints, barring collisions. 
    \item Inverse kinematics of serial structures are generally dependent on numerical calculation of forward kinematics. Using leg lengths as inputs results in a numerical approximation wrapped inside numerical approximation, which is not time efficient.
\end{itemize}

Assuming an inverse kinematic solution is obtained, verifying that the solution is optimal is a challenging task. The 24DOF overactuation of the robot leads to a large space of valid solutions for any desired end effector position. Initial work on choosing optimal positions to maximize the structural stiffness of an Assembler has been performed\cite{balaban_inverse_2018}, but was dependent on the assumption that the Assembler be confined to motion in two dimensions. The method proposed here extends this capability to three dimensions.

\section{Approximating Valid Serial-Parallel IK Solutions}
Na\"ive implementations of a nonlinear solver for the Assembler platform fail due to the range of possibilities for solving in a global space, requiring an initial condition. In order to hasten the solution finding process, it was necessary to generate an algorithm which could operate efficiently, and provide a valid, workable point which could serve as the initial condition for a constrained nonlinear solver.

\begin{figure}[b]
    \centering
    \includegraphics[width=.5\linewidth]{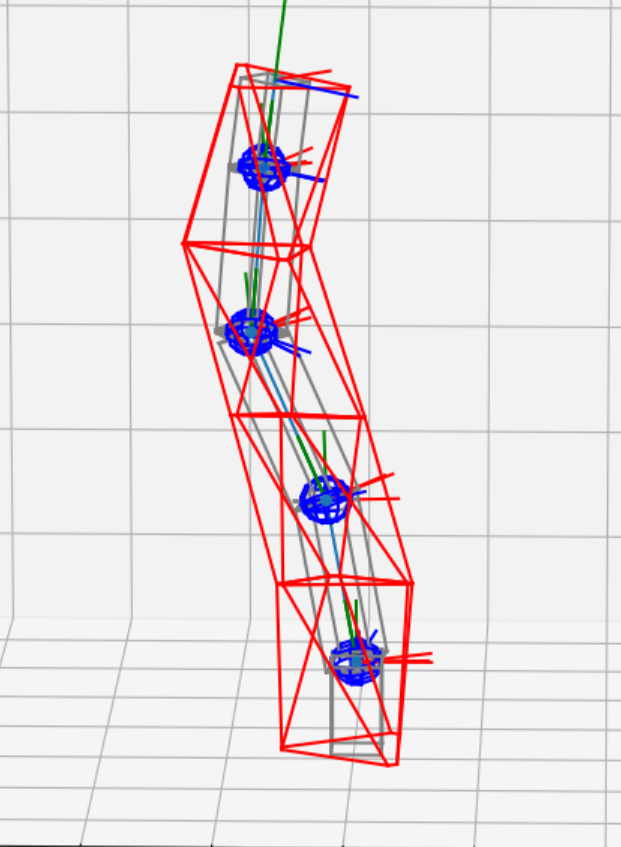}
    \caption{Simulated Assembler (red) with helper arm (gray) visible. The blue spheres represent the helper arm's triple joint clusters.}
    \label{fig:helperArm}
\end{figure}

The initial condition algorithm requires the creation of a "helper" serial arm shown in Fig. \ref{fig:helperArm}, which assists in the inverse kinematics of the stack. The helper arm consists of a 3n DOF serial arm, where n is the number of platforms in the stack. Each set of 3 joints on the helper arm are co-located together, and consist of revolute joints about the x, y, and z-axes. Each joint cluster is initially placed \emph{$d_0$} meters apart, where \emph{$d_i$} is the distance between the top and bottom plates of a Stewart platform component of the Assembler in the \emph{home pose}, which is defined to be the starting pose where all actuators are halfway extended.

The algorithm itself is recursive, and the general process is described here. It begins by solving constrained IK on the helper arm, and if it finds a valid solution to the goal pose, it places the Assemblers plates at the midpoints between the arm joints in a way to minimize angular deviation. If the arm does not find a valid configuration to the goal pose, it changes the value of \emph{$d_i$} and recurses, with \emph{$d_i$} scanning outward from the initial condition ($d_0$) with a user determined step size. This has the effect of changing the lengths of each link of the helper arm, approximating the internal extensions of each Stewart platform. If the algorithm hits a user defined recursion limit and no valid solution is found, it returns the Assembler home pose, and a failure flag.
Given that a successful position is found, and the platform plates are placed accordingly, the algorithm runs inverse kinematics on each constituent Stewart platform, appends their leg lengths into a 6xn matrix and returns that matrix as the output of the function. 

The purpose of this algorithm is to quickly generate a feasible initial condition for the Assembler to reach a goal position which does not violate any physical constraints. Other approaches to initializing a feasible solution may also be applied; further details of this and other options are beyond the scope of this paper.

\section{Application of Constrained Nonlinear Optimization in Serial-Parallel IK}
Using an initial condition generated from the algorithm in the previous section, it is possible use any nonlinear optimization method, such as Python's scipy-optimize\cite{2020SciPy-NMeth} package, to find a local minimum given a user-specified optimization metric.

\subsection{Constraints}
\subsubsection{Leg Lengths}
Each leg length, $\Li$, of a given platform must not exceed the maximum leg length $L_{max}$ or be smaller than the minimum leg length $L_{min}$ (\ref{legConstraint}).
\begin{equation}
    \label{legConstraint}
    L_{min} <= \Li <= L_{max}
\end{equation}

\subsubsection{Angular Deviation}
There is a maximum amount an actuator can deviate from its normal angle (defined as the angle created by the actuator's top and bottom anchor points and the bottom plate origin while all actuators are at 50\% extension), before the actuator snaps free of its housing and the robot fails. Assembler actuator angles from normal can be calculated by (\ref{angularDeviation}), where the denotation of rest indicates the position of the joint in the neutral pose.
\begin{equation}
   \label{angularDeviation}
    \theta_i = cos^{-1}\frac{(\spaceCoordsTopJoints_i-\spaceCoordsBotJoints_i)}{||\spaceCoordsTopJoints_i-\spaceCoordsBotJoints_i||}\bullet\frac{(\spaceCoordsTopJoints_i^{rest} - \spaceCoordsBotJoints_i^{rest})}{||\spaceCoordsTopJoints_i^{rest} - \spaceCoordsBotJoints_i^{rest}||}
\end{equation}

Because $\theta$ must be constrained to a maximum deviation angle $\theta_{max}$ this relationship must hold:

\begin{equation}
    \theta(\Li) <= \theta_{max}
\end{equation}

\subsubsection{Forces}
Each leg can tolerate a maximum compression force and a maximum tension force. These forces, denoted $F(\Li)$, are signed (positive for tension, negative for compression). The force calculation takes into account forces acting on the end effector of the platform, in addition to the masses of the plates, actuator motors, and actuator pistons. Calculation of the forces assumes the Assembler unit is a static truss, and utlizes wrench formulation as described in Modern Robotics \cite{lynch_modern_2017}. The maximum tension force in newtons can be denoted \emph{$F_{max}^T$}, while the maximum compression force can be denoted \emph{$F_{max}^C$}. The following relationship must hold (\ref{forceConstraint}).


\begin{equation}
    \label{forceConstraint}
    - F_{max}^T <= F(\Li) <= F_{max}^C
\end{equation}

\subsubsection{End Effector Position}
The initial condition to the equation $X_0$ will be determined from an existing algorithm that delivers a valid position for the platform stack where $\tm_{EE}$ must be equal to the goal position, $\tm_{Goal}$. In order to retain this validity while optimizing the midplates, the following constraint (\ref{posConstraint}) must be obeyed to ensure that none of the 6 end effector DOFs exceed predefined tolerances.

\begin{equation}
    \label{posConstraint}
    ||\taa_{{EE}_i}-\taa_{{Goal}_i}|| <= Tol_i 
\end{equation}
\subsubsection{Continuous Translation Enforcement}
In order to ensure that the optimization function does not attempt to place a plate that either intersects or is behind the previous plate, the z-translation between corresponding joints on any given pair of plates must be positive, enforcing the relationship:

\begin{equation}
    \spaceCoordsTopJoints_i^z > \spaceCoordsBotJoints_i^z
\end{equation}

\subsection{Optimization Function}


The optimization function (\ref{optimizationA}) is weighted between two terms. The first term selects the maximum leg force magnitude incurred throughout the structure. The second term calculates a measure of the average angular deviation of stewart platform plates from their resting poses. $p_i^{rest}$ specifically describes the resting pose of plate $i$ relative to the plate that precedes it. The weights $w_1$ and $w_2$ allow for the function to be tuned. For the purposes of this analysis, both were set to 1. 

\begin{equation}
     \lambda_i = \cos^{-1}(\frac{\trans_i-\trans_{i-1}}{||\trans_i-\trans_{i-1}||}\bullet \frac{\trans_{i}^{rest}-\trans_{i-1}}{||\trans_{i}^{rest}-\trans_{i-1}||})
\end{equation}
\begin{equation}
    \label{optimizationA}
\Omega(x) = w_1*\max(||(F(L_{1:26})||) + w_2\sqrt{\sum_{i=1}^4(\lambda_i)^2}
\end{equation}

%
The general form of the optimization function is to minimize $\Omega(x)$ where $x$ is a 1x18 vector consisting of the TAA poses in $O$ of SPs 1-3. SP4 is placed at the goal pose. Execution of the algorithm is subject to all constraints described in section 4.1, where $g_i(x) == True$.

\section{Simulation}
This section shows a representative set of poses that a 4-unit Assembler platform might need to optimize for during the course of normal operation in manipulating objects. In each pose, a given mass of $5kg$ is suspended $0.2m$ from the end effector in the end effector's Z local frame, while plates are defined to  weigh $0.2kg$, and actuators $0.14kg$, with the actuator shaft weighing $0.04kg$ and the actuator motor weighing $0.1kg$. Gravity in the simulation is at $9.81m/s^2$. Forces are measured in Newtons ($N$), with a negative force indicating compression of the actuator. For the purposes of the simulation, elements such as electronics and ball joints are massless, and the center of mass of actuators is perfectly in line with the load path. 

\subsection{Representative Poses}

\subsubsection{Horizontal Translations}
\begin{figure}[t]
    \centering
    \includegraphics[width=.7\linewidth]{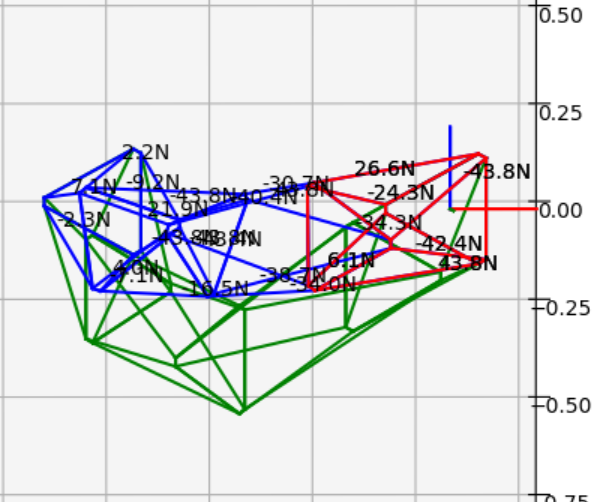}
    \caption{0.8m Horizontal Translation, Top View. Green represents the initial condition. The blue and red Assembler represents the optimized pose.}
    \label{fig:TranslationTop}
\end{figure}

\begin{figure}[t]
    \centering
    \includegraphics[width=.7\linewidth]{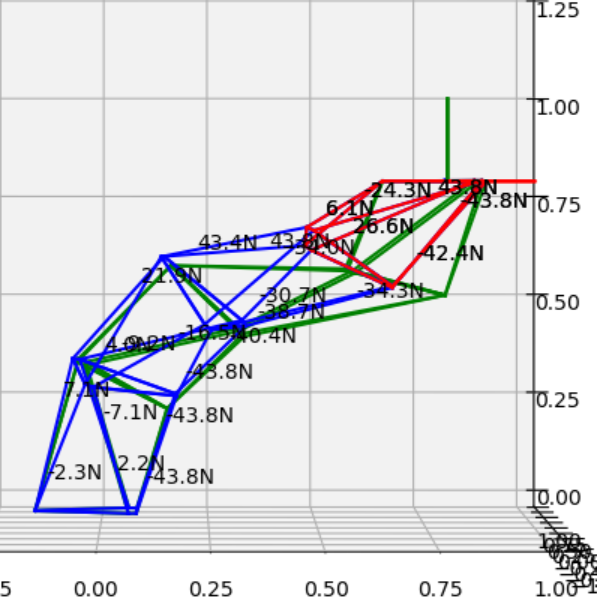}
    \caption{0.8m Horizontal Translation, Side View}
    \label{fig:TranslationSide}
\end{figure}

In most cases for horizontal translation of the end-effector, the initial condition algorithm selects a pose fairly close to optimal. An uncharacteristically suboptimal position was intentionally selected as the initial condition for this optimization. Looking at a side view seen in Fig. \ref{fig:TranslationSide}, the poses are fairly close together, with the green pose (the initial condition) presenting a slightly sharper internal angle between platforms 3 and 4. The difference between the two poses becomes more apparent when viewed from the top view shown in Fig. \ref{fig:TranslationTop}. The initial condition pose has a large swing out to the side; an artifact of how the pose is generated using a helper arm. 

Table \ref{table:Translation} shows the initial and final forces on the actuators in newtons. The force table was included for this pose due to more fully express the optimization the platform followed. The maximum absolute force of the initial pose is a compression load of $145 N$, incurred on the first actuator on platform 1, directly beneath the overhang. The actuator is at a minimum extension to accommodate the lateral expansion shown in Fig. \ref{fig:TranslationTop}. Optimization of the platform was able to reduce the maximum force from $145 N$ to $43.8 N$, still on the same actuator. This massive reduction of force was made possible by shifting the centers of mass of the Assembler plates inline between the base and the end effector. The mean magnitude of forces on all actuators was reduced from $50N$ to $27N$, mostly in part to outliers such as the actuator described above. Additionally, the sharpness of the internal angles of the plates was reduced as well, which is a result of the secondary part of the objective function, attempting to reduce angular plate deviation locally between platforms.

\begin{table}[H]
\addtolength{\tabcolsep}{-5pt}
\caption{0.8m Horizontal Translation Initial and Final Forces (N)}
\label{table:Translation}
\begin{minipage}{.47\columnwidth}
\centering
\begin{tabular}{ c  c  c  c }
\hline
\toprule
SP1 & SP2 & SP3 & SP4\\
\midrule
-145.609 & -17.601 & 12.723 & -44.693\\
69.279 & -89.122 & -139.754 & -23.686\\
18.444 & 62.559 & -12.84 & -1.134\\
15.428 & 54.751 & 60.392 & -29.233\\
-46.774 & -12.029 & 122.407 & 22.693\\
27.774 & -64.647 & -66.273 & 36.504\\
\bottomrule
\end{tabular}
\end{minipage}
\begin{minipage}{.47\columnwidth}
\centering
\begin{tabular}{ c  c  c  c }
\hline
\toprule
SP1 & SP2 & SP3 & SP4\\
\midrule
-43.842 & -43.788 & -38.748 & -42.436\\
-43.841 & -40.403 & -34.295 & -43.791\\
2.208 & -9.193 & -30.702 & 26.579\\
7.062 & 21.92 & 43.842 & -24.276\\
-2.327 & 4.039 & 43.435 & 6.074\\
-7.125 & -16.472 & -33.979 & 43.841\\
\bottomrule
\end{tabular}
\end{minipage}

\end{table}

\subsubsection{90 Degree Horizontal Translations}
\begin{figure}[b]
    \centering
    \includegraphics[width=.7\linewidth]{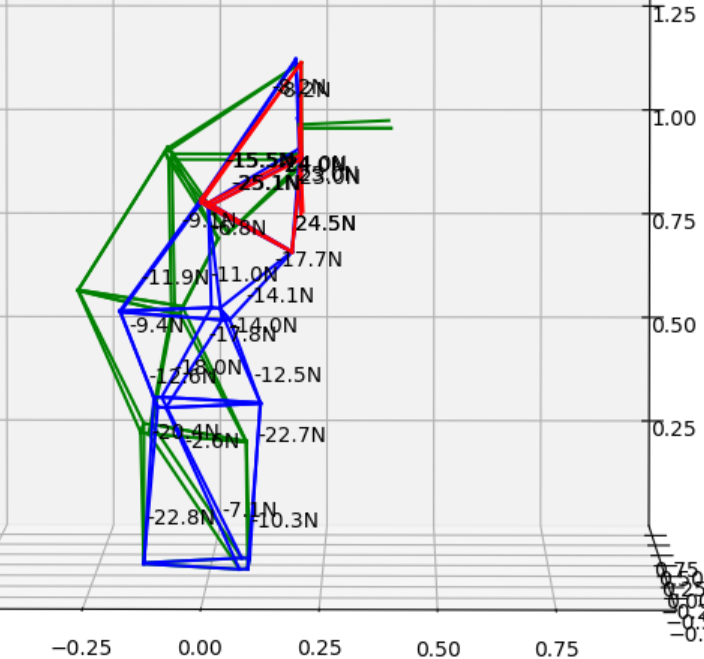}
    \caption{Assembler Positioned with End Effector 90 Degrees Relative to the Base Plate}
    \label{fig:90DegSide}
\end{figure}

The Assembler may also be required to place items onto fixtures which are perpendicular to the mountings of the robot. This pose represents the beginning of a linear translation forward, where the end effector will be kept at an angle of 90 degrees relative to the base plate. With the end effector of the robot only 0.2m in front of the origin point as shown in Fig. \ref{fig:90DegSide}, the robot was able to collapse its platforms towards the base, slightly increasing the interior angles on the final two platforms, but decreasing the overall forces incurred by the actuators. The maximum force drops by 50\% from $40.9 N$ to $20.2 N$, while the mean force drops from $20.2N$ to $15N$.

\subsection{Inchworming Locomotion}
\begin{figure}[t]
    \centering
    \includegraphics[width=.7\linewidth]{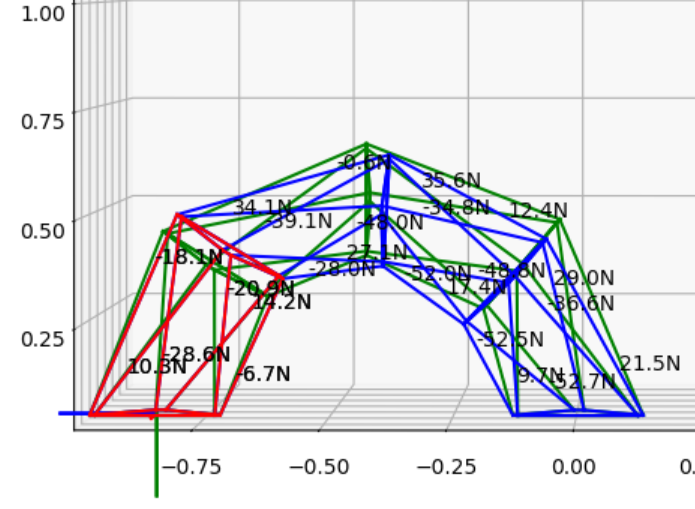}
    \caption{Assembler Inchworming Pre-Contact Force Comparison}
    \label{fig:BridgePose}
\end{figure}

In situations where the Assembler unit is not being moved about in the global work space by a long reach manipulator such as the LSMS, its only available locomotion options are to either roll, or to inchworm itself across the surface of its environment. This case describes such a motion, where the Assembler reaches down to the floor. This case considers load optimization of the Assembler whilst it is still supported at a single point, immediately prior to contact with its end-effector. The comparison of initial and force optimized poses is in Fig. \ref{fig:BridgePose}.
\begin{table}[t]
\addtolength{\tabcolsep}{-5pt}
\caption{Inchworming Locomotion Initial and Final Forces (N)}
\label{table:inchworm}
\begin{minipage}{.47\columnwidth}
\centering
\begin{tabular}{ c  c  c  c }
\hline
\toprule
SP1 & SP2 & SP3 & SP4\\
\midrule
22.77 & -57.781 & 8.631 & -21.559\\
-45.112 & 31.048 & -28.067 & 3.702\\
13.999 & 16.384 & 6.369 & -12.622\\
49.68 & -27.722 & 19.247 & -24.284\\
-80.622 & 3.444 & -44.769 & 8.622\\
-34.229 & -36.209 & -21.419 & -4.913\\
\bottomrule
\end{tabular}
\end{minipage}
\begin{minipage}{.47\columnwidth}
\centering
\begin{tabular}{ c  c  c  c }
\hline
\toprule
SP1 & SP2 & SP3 & SP4\\
\midrule
9.69 & -48.826 & 27.146 & -20.886\\
-36.586 & 35.615 & -39.13 & 10.33\\
21.466 & 12.353 & -0.649 & -18.066\\
28.976 & -34.782 & 34.096 & -28.573\\
-52.682 & 17.429 & -48.02 & 14.188\\
-52.502 & -51.991 & -27.986 & -6.709\\
\bottomrule
\end{tabular}
\end{minipage}

\end{table}
In this scenario, the mean absolute force was not actually reduced, but the maximum force was. In Table \ref{table:inchworm}, included due to the interesting nature of this optimization, the maximum applied force was reduced from a magnitude of 80.6N to $52.7N$, while the mean force increased from $25.9N$ to $28.3N$. The reason for this phenomenon was that in the initial, evenly distributed pose, the largest forces were all concentrated in the bottom platform, while latter platforms experienced lessened forces. The optimizer pulled the plates back towards the origin as much as possible while staying within angular constraints, reducing the maximum forces incurred on the bottom plate while overall forces became more tightly clustered. The greatest magnitude of force changes occurred on the bottom platform.

\begin{figure}[b]
    \centering
    \includegraphics[width=.75\linewidth]{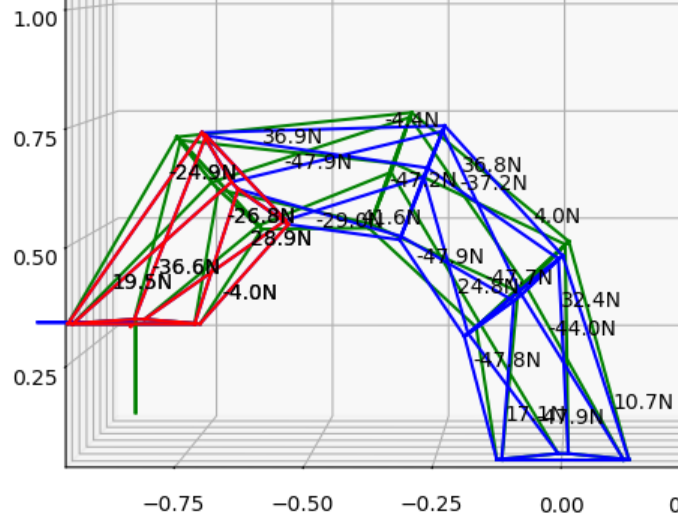}
    \caption{Assembler Vertical Lift}
    \label{fig:VerticalLift}
\end{figure}
\subsection{Lifting Objects Vertically from the Floor}
Superficially similar to the inchworming position, this pose represents a use case where the Assembler may be required to pick an object off of the floor nearby before using it in a task. This pose assumes that the Assembler must translate its end effector directly downward to achieve a connection to the desired object. Like the inchworming position, the maximum force of $73.1N$ is reduced to $47.9N$ while the mean force of $26.5N$ is increased to $31N$. Plate positions again shrink closer to the origin in an effort to shift the center of mass away from the end effector. This reduces strain on the bottom plate, but also on the compression forces placed on the inside actuators.

\subsection{Orientation Transition}
\begin{figure}[t]
    \centering
    \includegraphics[width=.75\linewidth]{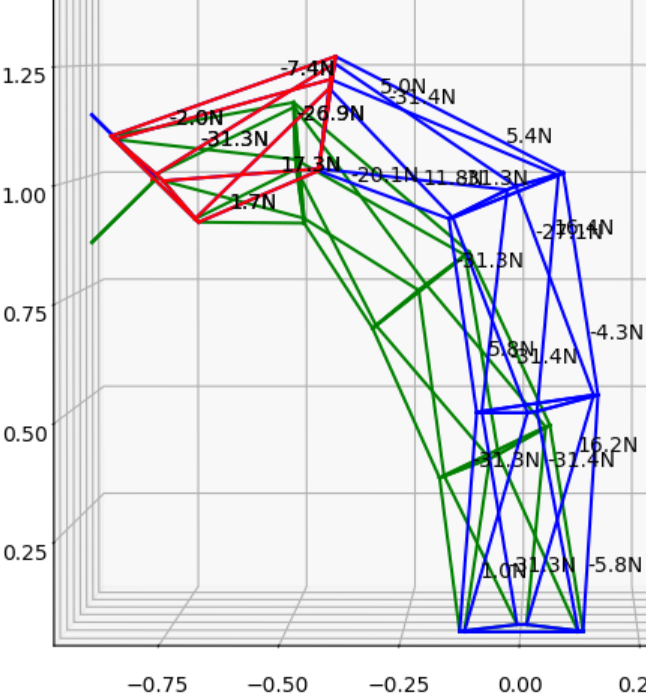}
    \caption{Orientation Transition}
    \label{fig:OrientationTransition}
\end{figure}

In the process of lifting an object, the Assembler may have to transition the object to different orientations. The pose shown in Fig. \ref{fig:OrientationTransition} demonstrates one such case, while also affording a pose with excellent room for optimization, as shown. The optimized pose arches backward behind the base position of the Assembler to shift the center of mass behind the origin and better distribute loads on the platforms. The maximum force is cut in half from a force of $66.2N$ in platform 3 to $31.3N$ in platform 1. In this case, the mean force is also reduced from $24.5N$ to $17.7N$ as the Assembler is able to better distribute loads across its base pose.


\section{Future Work}
The optimization work accomplished here will enable the Assembler units to sustain heavier loads per actuator compared to average valid poses. This capability will improve the usefulness of the Assembler for manipulation tasks, and open the door to a variety of new research areas. 

\begin{figure}[t]
    \centering
    \includegraphics[width=.78\linewidth]{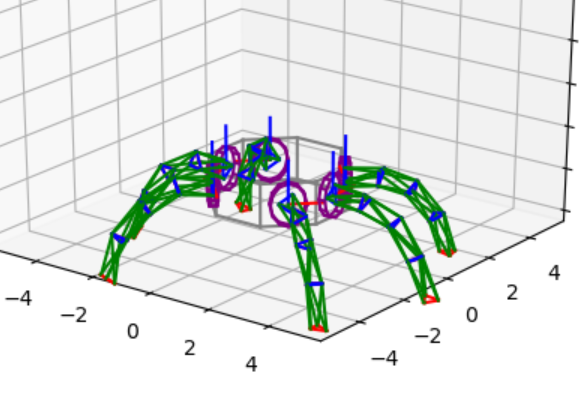}
    \caption{Hexapod Platform Made of Assembler Stacks}
    \label{fig:Hexapod}
\end{figure}

Physical prototypes of truss-like robots such as the Assembler system will be evaluated for use in surface and orbital environments via demonstration of multi-agent autonomous assembly. 
Multiple Assembler units could join together in macro-configurations such as the hexapod-type robot shown in Figure \ref{fig:Hexapod}, which would be able to carry heavy payloads beyond the capacity of any single Assembler unit over great distances, improved by the force optimization described here. This configuration would also allow for the carrying of additional computational power in the central core area, supporting assembly tasks by coordinating nearby robots.

The force optimization function itself can be improved by increasing the computational speed of the function, or optimizing by additional optional metrics such as structural stiffness of the Assembler or minimum required movement to get from position to position. It is worth looking into motion planning as a function of optimimal positions, and developing more accurate force simulation, such as adding support for multi-point force calculations. 

\section{Conclusion}


The range of applications for modular truss-like robots encompasses both orbital and terrestrial domains. On-orbit, cooperative systems can build and maintain persistent infrastructure and use their strength and precision to form the backbone of reconfigurable structures, such as modular telescopes. In terrestrial applications, Assembler macro-configurations could explore hazardous areas, and establish initial work sites where permanent infrastructure could be later erected by multi-agent robotic teams. 

The ability to now generate force-optimized actuator configurations given a desired end-effector configuration not only opens the door to higher fidelity simulation, but increases the capability of the robot to safely manipulate useful payloads. Predictable positioning of internal components of the robots will drive more complete motion planning algorithms, while results from hardware tests informed by this new kinematic model will allow for the development of an eventual demonstration consisting of a multi-agent robotic team erecting a simple structure such as a mock habitat or backbone truss. 

\addtolength{\textheight}{-12cm}   




\section*{ACKNOWLEDGMENTS}

The authors thank Samuel Schoedel and Alex Fuller, undergraduates at Virginia Tech, for their efforts in hardware prototyping.


\bibliographystyle{IEEEtran}
\bibliography{CombinedBibTeX}

\end{document}